\documentclass[10pt,twocolumn,letterpaper]{article}

\usepackage{cvpr}              

\usepackage{multirow}

\definecolor{cvprblue}{rgb}{0.21,0.49,0.74}
\usepackage[pagebackref,breaklinks,colorlinks,allcolors=cvprblue]{hyperref}

\def\paperID{16748} 
\def\confName{CVPR}
\def\confYear{2025}

\title{SASep: Saliency-Aware Structured Separation of Geometry and Feature \\ for Open Set Learning on Point Clouds}

\author{
    \hspace{-8pt}Jinfeng Xu\textsuperscript{1}\hspace{7pt} 
    Xianzhi Li\textsuperscript{1,2}\hspace{7pt} 
    Yuan Tang\textsuperscript{1}\hspace{7pt} 
    Xu Han\textsuperscript{1}\hspace{7pt} 
    Qiao Yu\textsuperscript{1}\hspace{7pt} 
    Yixue Hao\textsuperscript{1}\hspace{7pt} 
    Long Hu\textsuperscript{1}\thanks{Corresponding author.}\hspace{7pt} 
    Min Chen\textsuperscript{3,4}\\
    \hspace{-8pt}\textsuperscript{1}Huazhong University of Science and Technology\hspace{10pt}
    \textsuperscript{2}Guangdong Intelligent Robotics Institute\\
    \hspace{-8pt}\textsuperscript{3}South China University of Technology\hspace{100pt}
    \textsuperscript{4}Pazhou Laboratory\\
    \hspace{-8pt}{\tt\small \{jinfengx, xzli, yuan\_tang, xhanxu, qiaoyu\_epic, yixuehao, hulong, minchen\}@hust.edu.cn}
}

\begin{document}
\maketitle
\begin{abstract}

Recent advancements in deep learning have greatly enhanced 3D object recognition, but most models are limited to closed-set scenarios, unable to handle unknown samples in real-world applications. 
Open-set recognition (OSR) addresses this limitation by enabling models to both classify known classes and identify novel classes.
However, current OSR methods rely on global features to differentiate known and unknown classes, treating the entire object uniformly and overlooking the varying semantic importance of its different parts.
To address this gap, we propose \textbf{S}alience-\textbf{A}ware Structured \textbf{Sep}aration (SASep), which includes (i) a tunable semantic decomposition (TSD) module to semantically decompose objects into important and unimportant parts, (ii) a geometric synthesis strategy (GSS) to generate pseudo-unknown objects by combining these unimportant parts, and (iii) a synth-aided margin separation (SMS) module to enhance feature-level separation by expanding the feature distributions between classes.
Together, these components improve both geometric and feature representations, enhancing the model’s ability to effectively distinguish known and unknown classes.
Experimental results show that SASep achieves superior performance in 3D OSR, outperforming existing state-of-the-art methods. 
The codes are available at \href{https://github.com/JinfengX/SASep}{https://github.com/JinfengX/SASep}.

\end{abstract}    
\section{Introduction}
\label{sec: intro}

\begin{figure}[!t]
  \centering
  \includegraphics[width=\linewidth]{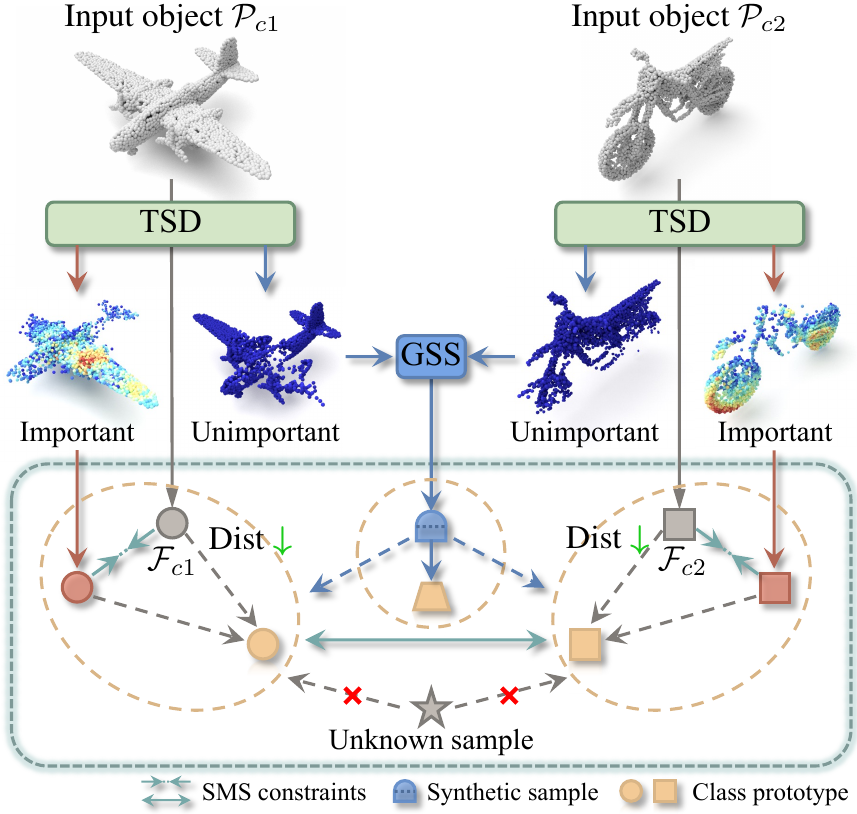}
    \vspace{-0.5cm}
   \caption{ 
   Given input objects, the TSD module semantically decomposes them into important and unimportant parts based on each point's semantic relevance to classification.
   The important parts enhance feature learning for known classes, while the unimportant parts are mixed in GSS to generate pseudo-unknown samples. 
   The SMS module then constrains the features of objects and their decomposed parts to improve inter-class separation and intra-class compactness.
   }
   \label{fig: teaser}
   \vspace{-0.2cm}
\end{figure}

In recent years, deep learning has significantly advanced 3D object recognition, enabling its application in diverse areas such as autonomous driving, robotics, augmented/virtual reality, \etc.
Most existing 3D recognition models~\cite{Qi_2017_CVPR, Zhao_2021_ICCV, Wu_2024_CVPR} are designed for closed-set scenarios, where all test classes are assumed to have been seen during training.
However, real-world applications usually present dynamic environments where unknown or unseen samples outside the set of known classes may be encountered.
These out-of-distribution samples pose a significant challenge, as the closed-set models are unable to handle these unknown samples appropriately, often misclassifying them as one of the predefined known classes.
To address this challenge, open-set recognition (OSR) task has gained increasing attention recently, as it aims to extend the capabilities of closed-set models by enabling them to classify known classes and identify unknown ones simultaneously.

Early 2D OSR approaches~\cite{hendrycks2017a, pmlr-v162-hendrycks22a, liang2018enhancing} aim to distinguish between known and unknown classes by using discriminative functions to compute confidence scores from network predictions for each input.
Samples are identified as unknown when their confidence scores are low.
Thus, \emph{designing a robust open-set solver hinges on ensuring the confidence score is discriminative}, meaning scores for known classes should clearly exceed those for unknown classes.

Although recent 2D OSR works~\cite{9521769, NEURIPS2021_01894d6f, Wang_Mu_Zhu_Hu_2024} have achieved significant progress in the 2D area, research on OSR for 3D data remains relatively underdeveloped.
Unlike 2D OSR, the unique spatial characteristics of 3D data limit the performance of directly adapting 2D OSR methods to 3D, as they fall short of leveraging 3D-specific structural and topological information.
Recent works~\cite{9506795, BHARDWAJ2021172, weng2024partcom} propose to address the 3D OSR task from the perspective of geometric reconstruction, feature embeddings, and sample synthesis.
However, these approaches, similar to many 2D OSR methods, primarily focus on assigning confidence scores based on whole-object features, often overlooking the fact that \textbf{different parts of an object contribute differently to classification}.
For example, specific regions, such as the wheels of a car or the wings of an airplane, carry essential features that are more informative for classification than other parts.

Based on this observation, we propose a novel method for 3D open-set recognition on point clouds, named \textbf{S}alience-\textbf{A}ware Structured \textbf{Sep}aration (SASep), which tackles the open-set point cloud learning task by highlighting the semantically important geometric structures of objects.
As illustrated in~\cref{fig: teaser}, SASep first decomposes the input objects into important and unimportant parts with a tunable semantic decomposition (TSD) module based on the relevance of each region to the class semantics.
Then, we augment the feature learning of known classes by incorporating important parts into the training dataset.
For the unimportant parts, we design a geometric synthesis strategy (GSS) to generate synthetic samples, serving as pseudo-unknown objects to simulate geometric structures of unknown classes. 
Furthermore, we refine the feature space with synthe-aided margin separation module (SMS) by applying constraints that reduce the feature distance between the object and its important parts, while increasing the feature distance between the object and those from other classes.
This process enhances the feature representations of known classes, thus reducing the likelihood of misclassifying unknown class samples as known classes.
The experimental results demonstrate that our proposed method outperforms existing OSR approaches, achieving state-of-the-art (SOTA) performance in 3D open-set recognition.
Overall, our contributions can be summarized as follows:
\begin{itemize}
    \item We introduce a novel approach for 3D open-set recognition on point clouds, called Salience-Aware Structured Separation (SASep), which enhances the distinction between known and unknown classes by utilizing the semantically decomposed parts of objects.
    \item We propose a novel semantic decomposition method TSD that decomposes an object into semantically important and unimportant regions, enabling the model to utilize structural information from different parts of the objects.
    \item We design GSS and SMS to utilize the decomposed parts of TSD to explore feature distribution of the unknown class and enhance feature distinction of the known classes, respectively.
\end{itemize}
\section{Related work}
\label{sec:related work}

\subsection{Closed-set point cloud learning}

Closed-set point cloud learning has made significant progress, focusing on accurate classification and segmentation of 3D objects from predefined categories.
Existing approaches can be broadly categorized into view-based, voxel-based, and point-based methods.
View-based methods~\cite{Chen_2017_CVPR, Lang_2019_CVPR, Tatarchenko_2018_CVPR, Su_2015_ICCV} classify 3D shapes by using 2D CNN to extract features from the projected 2D images of input.
Voxel-based methods~\cite{7353481, Choy_2019_CVPR, Graham_2018_CVPR, Lei_2019_CVPR, 10.1145/3072959.3073608} process point clouds by dividing 3D space into voxel grids, allowing 3D CNNs to learn the spatial features.
Pioneering point-based methods~\cite{Qi_2017_CVPR} use Multi-Layer Perceptrons (MLPs) with symmetric functions to directly process raw point cloud data.
The following point-based research~\cite{qi2017pointnet++, wang2019dynamic, 10.1145/3664647.3681173, ma2022rethinking, Lai_2022_CVPR} further improved the performance by enhancing the models' capacity to aggregate the local and global features. 
While these methods have achieved significant success in recognizing known classes, they face notable limitations when dealing with unseen classes, as the models usually tend to misclassify the unknown objects as one of the predefined known classes.
This limitation restricts the models' ability to generalize to real-world applications, where novel or unknown classes may be encountered during inference.

\begin{figure*}[!ht]
  \centering
  \includegraphics[width=\textwidth]{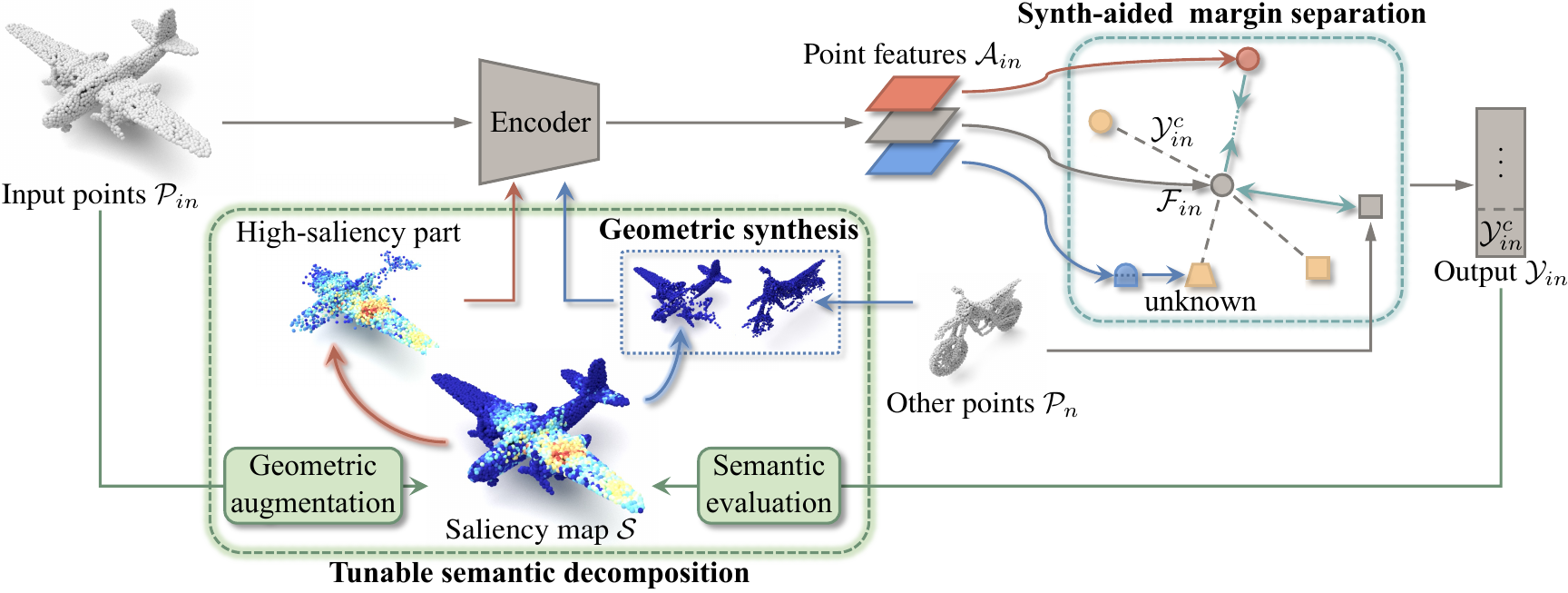}
    \vspace{-0.5cm}
   \caption{Overall architecture of our SASep. The architecture includes a forward process where the input point cloud $\mathcal{P}_{in}$ is encoded with backbone, followed by a backward process utilizing the tunable semantic decomposition module to decompose $\mathcal{P}_{in}$ into high- and low-saliency parts. Then, the high-saliency parts are used in the synth-aided margin separation module for separating feature distribution; the low-saliency parts are combined for synthetic objects with a geometric synthesis strategy.}
   \label{fig: architecture}
   \vspace{-0.2cm}
\end{figure*}

\subsection{Open-set 2D recognition}

To address the limitation of closed-set learning, open-set recognition (OSR) enables models to classify known objects and identify unknown or novel classes during inference.
The straightforward approaches~\cite{liang2018enhancing, Wang_2021_ICCV, NEURIPS2021_01894d6f, NEURIPS2021_063e26c6} utilize discriminative functions to evaluate confidence scores for identifying unknown classes.
MSP~\cite{hendrycks2017a} established a robust baseline by using Maximum Softmax Probability to identify samples as unknown when they exhibit low maximum softmax scores.
Similarly, Maximum Logit Score (MLS)~\cite{pmlr-v162-hendrycks22a} directly uses the highest logit score as a confidence measure.
Beyond these discriminative-based approaches, reconstruction-based methods~\cite{Kong_2021_ICCV, du2022towards, 8461700} identify unknown objects by comparing the original inputs with their reconstructions from the generative models.
Recent research in 2D OSR aims to learn feature embeddings that form compact clusters for known classes.
For instance, ARPL~\cite{9521769} introduces reciprocal points in the embedding space to minimize the overlap of known and unknown distributions.
Fontanel~\etal~\cite{Fontanel_2021_CVPR} proposed a method that uses learnable class prototypes to detect anomalies by measuring the cosine distance of inputs from learned class prototypes.
Furthermore, several studies~\cite{Neal_2018_ECCV, du2022vos, Liu_2023_CVPR} have investigated using synthetic samples to expose potential unknown classes to the model, thereby enhancing the model's generalization on samples out of the training dataset.

Though 2D OSR has seen significant advancements, applying these methods to 3D domains presents limitations, particularly in leveraging the unique spatial and geometric complexities within 3D data.
Thus, directly shifting a 2D OSR method to 3D data often fails to effectively distinguish known and unknown samples.

\subsection{Open-set 3D recognition}

In contrast to the widely explored 2D OSR, research on 3D OSR remains relatively underdeveloped.
Masuda~\etal~\cite{9506795} developed a reconstruction-based method to obtain anomaly scores by measuring reconstruction errors between model input and output.
To enhance the learning of distribution for unknown classes, Bhardwaj~\etal~\cite{BHARDWAJ2021172} proposed generating pseudo-unknown samples by mixing objects from known classes.
To further refine the synthesis strategy, Lee~\etal~\cite{Lee_2021_CVPR} and Hong~\etal~\cite{hong2022pointcam} designed more sophisticated algorithms to synthesize high-quality pseudo unknown samples, considering the local geometric structure of objects.
More recently, PartCom~\cite{weng2024partcom} introduced a novel part codebook to learn 3D object representations by capturing distinct part compositions across classes.
Similarly, Rabino~\etal~\cite{10684168} introduced a training-free method leveraging part-based reasoning.
Building on the work of 2D and 3D OSR, 3DOS~\cite{NEURIPS2022_85b6841e} provided a comprehensive benchmark for 3D open-set learning, revealing the strengths and limitations of existing approaches in the 3D open-set context.

Existing open-set recognition methods focus on distinguishing known and unknown objects or features as a whole, while overlooking the essential semantic information carried by different parts in an object.
For example, the wheels of vehicles often exhibit more significant class-relevant information than the other parts.
Humans can recognize objects even with a partial view of such recognizable geometric or topological structures.
Building on this observation, we develop a method to address the challenges of open-set learning by focusing on the semantically important parts within 3D objects.
Our proposed method strengthens the association between class semantics and object partial structures, ultimately improving the accuracy and robustness of 3D open-set recognition.

\section{Method}
\label{sec: method}

\subsection{Architecture overview}
\label{subsec: overview}

As introduced in~\cref{sec: intro}, we propose the Saliency-Aware Separation (SASep) method to address the limitations of current methods, which overlook the different contributions of object parts to classification.
Figure~\ref{fig: architecture} illustrates the architecture overview of our SASep.
Building upon the basic pipeline of using an encoder to lift the input point cloud $P_{in}$ into the global feature $\mathcal{F}_{in}$, the closed-set model output $\mathcal{Y}_{in} \in \mathbb{R}^C$ can be formulated as the softmax of $g(\mathcal{F}_{in})$, where $g(\cdot)$ maps the global feature to a set of logits of $C$ known classes.
Rather than using the commonly adopted linear layer to implement $g(\cdot)$, we formulate $\mathcal{Y}_{in}$ by employing a prototype-based method~\cite{Fontanel_2021_CVPR} (see~\cref{eq: cosine_sim}), aiming to help capture important geometric characteristics of each class.
Then, we calculate the confidence score $\mathcal{Q}$ by extracting the maximum logit score (MLS) from $\mathcal{Y}_{in}$.
Objects with a $\mathcal{Q}$ value below a predefined threshold are classified as belonging to the unknown class.
%

To enhance this basic pipeline, we first design a tunable semantic (TSD) module (marked by a green dash box) to evaluate the semantic saliency of each point, thereby dividing objects into important (high-saliency) and unimportant (low-saliency) parts based on the level of semantic saliency.
For the high-saliency parts, we directly treat them as new instances of the same known class as the original objects and re-input them into the feature encoder to enhance the network’s learning ability for this category.
Note that, for the low-saliency parts that carry less known-class-specific information, we do not discard them.
Instead, we develop a geometric synthesis strategy (GSS) to create unknown objects by combining low-saliency parts from various known objects.
The rationale behind GSS lies in the fact that unknown- and known-class objects are likely to share similar local geometric structures. By re-inputting these synthetic objects into the network, the network's discriminative ability for similar regions can be enhanced.

Please note that the above designs are processed at the geometric level.
Given the difficulty in accurately recognizing unknown classes, we take a further step by designing a synth-aided margin separation (SMS) module (marked by a blue dash box) that operates $\mathcal{F}_{in}$ at the feature level, with the key idea of enlarging the feature distribution margins between classes.
In the following sections, we will provide detailed information about each of our designs.

\subsection{Tunable semantic decomposition module}
\label{subsec: TSD_module}

\begin{figure}[t]
  \centering
  \includegraphics[width=\linewidth]{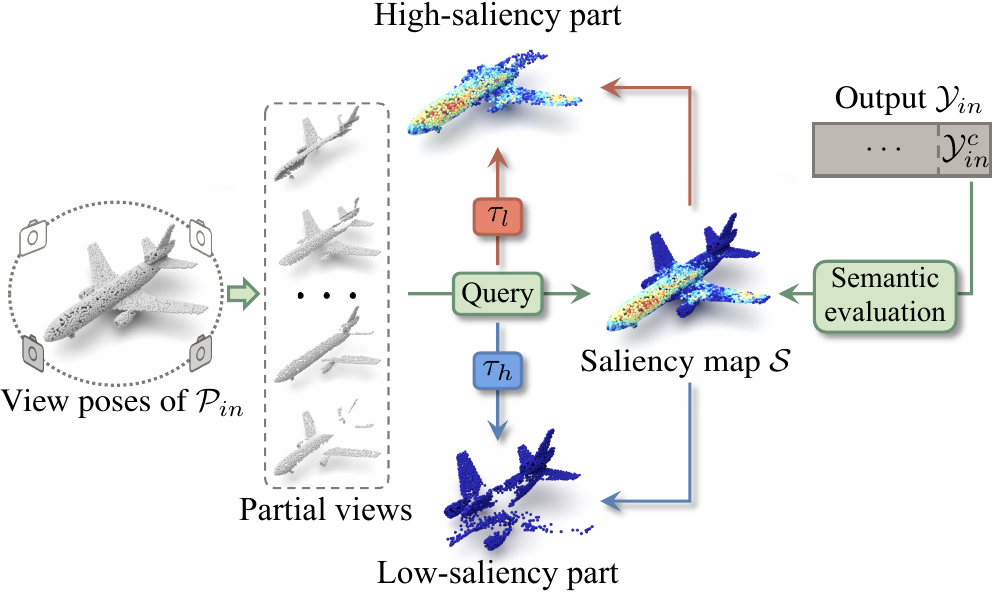}

    \vspace{-0.2cm}
   \caption{
   Illustration of our proposed tunable semantic decomposition (TSD) module.
   The TSD module decomposes the input objects into high- and low-saliency parts using the semantically evaluated saliency map $\mathcal{S}$.
   It also incorporates random partial views to enhance geometric information in a tunable manner.
   }
   \label{fig: TSD_module}
   \vspace{-0.3cm}
\end{figure}

The tunable semantic decomposition (TSD) module aims to decompose the objects into high-saliency parts and low-saliency parts, based on each point's contribution to classification predictions.
The pipeline of our proposed TSD module is illustrated in~\cref{fig: TSD_module}.
For the input point cloud $\mathcal{P}_{in}$, the model extracts its per-point features $\mathcal{A}$ to predict the output $\mathcal{Y}_{in}$.
Since $\mathcal{Y}_{in}$ reflects the class probabilities for $\mathcal{P}_{in}$, we aim to estimate the contribution of each point to the corresponding object class probability in $\mathcal{Y}_{in}$.
To achieve this goal, we adopt the Grad-CAM approach~\cite{Selvaraju_2017_ICCV} to perform semantic evaluation, which aggregates each channel of the point-wise feature $\mathcal{A}$ with associated gradient information to generate a saliency map $\mathcal{S}$:
\begin{equation}
    \mathcal{S} = \mathit{ReLU} \left( \sum_k \bigg( \frac{1}{N} \sum_i \frac{\partial \mathcal{Y}_{in}^{c}}{\partial \mathcal{A}^k_i}   \bigg) \mathcal{A}^k \right),
\end{equation}
where $\mathcal{A}^k_i$ denotes the $k$-th channel of $\mathcal{A}$ corresponding to the $i$-th point and $N$ is the number of points.
Generally, a point with a higher saliency score in $\mathcal{S}$ indicates that the point has a greater influence on the classification result, \ie, it is more important.
Then, we sort $\mathcal{S}$ in ascending order, designating the first $\left\lfloor N / M \right\rfloor$ points as the low-saliency part, while the remaining points are marked as the high-saliency part. Here, $M$ is a parameter that determines the number of low-saliency parts used in the geometric synthesis strategy.
Because the saliency map $\mathcal{S}$ focuses on evaluating the semantic relevance, the saliency-based decomposition tends to generate relatively fixed decomposed parts.
However, diverse scenarios, such as occlusion, noise, or corrupted data, are common in real-world point cloud data, which makes it difficult for the saliency-based decomposition method to effectively capture the key parts.
To improve the robustness and generalization of the model in such challenging conditions, we introduce a geometric augmentation strategy to the TSD module.
Specifically, by estimating visibility from view poses with random radii around the input points~\cite{mehra2010visibility}, we obtain the cropped parts of objects from these partial views.
For each partial view, the saliency scores of each point are retrieved from $\mathcal{S}$
Unlike simply augmenting the dataset by appending additional partial view data, we replace low- and high-saliency parts with partial views that have overall scores above $\tau_h$ and below $\tau_l$ ($\tau_h > \tau_l$), respectively.
By fine-tuning $\tau_h$ and $\tau_l$, the TSD module can flexibly shift the decomposition strategy from semantically focused (when $\tau_h$ is high and $\tau_l$ is low) to geometrically focused (when $\tau_h$ is low and $\tau_l$ is high).
This adaptability allows the TSD module to generate different samples that capture both semantic importance and geometric diversity.

\subsection{Geometric synthesis strategy}
\label{subsec: GSS}
To address the scarcity of unknown class data during training, the geometric synthesis strategy (GSS) synthesizes pseudo-unknown samples by mixing multiple low-saliency parts generated from the TSD module, thereby expanding the dataset to explore potential combinations of local structures that unknown classes may encompass.
Unlike methods that synthesize data using random parts of known objects, which risk introducing class-specific geometric structures, GSS limits impact on classification by leveraging less influential low-saliency parts.

Specifically, the low-saliency parts undergo standard transformations, followed by a mixing operation where $M$ randomly selected parts from different objects are combined.
The mixed objects are then encoded to generate outputs $\mathcal{Y}_s$, which are supervised using pseudo-labels $\hat{\mathcal{Y}_s}$ set to $C$+1.
To reduce overconfidence, label smoothing with base smooth weight $\epsilon$ is applied to $\hat{\mathcal{Y}_s}$.
Additional smooth weight $\epsilon_h$ is also assigned to the involved object classes, ensuring the mixed objects retain characteristics of their originals.
Assuming that the mixed parts are from classes $\mathcal{K}_s = \{ c_1, \cdots, c_n \}$, we define $\hat{\mathcal{Y}_s}$ as:
\begin{equation}
    \hat{\mathcal{Y}_s^i} = 
    \begin{cases}
        1 - \epsilon - \epsilon_h, & \mathit{if}\ i = C+1\\
        \epsilon / C,  & \mathit{if}\ i \in \mathcal{K}\ \mathit{and}\ i \notin \mathcal{K}_s\\
        \epsilon / C + \epsilon_h M_{c_i} / M, & \mathit{if}\ i \in \mathcal{K}_s,
    \end{cases}
\end{equation}
where $M_{c_i}$ is the number of mixed objects belonging to $c_i$.
Finally, the loss of GSS is formulated as:
\begin{equation}
\label{eq: gss_loss}
    \mathcal{L}_s = - \sum_{i=1}^{C+1} \hat{\mathcal{Y}_s^i} \log \frac{e^{\mathcal{Y}_{s}^i}}{\sum_{j=1}^{C+1} e^{\mathcal{Y}_{s}^j}},
\end{equation}
where $\mathcal{Y}_{s}^i$ denotes the $i$-th class probability of $\mathcal{Y}_s$.

\subsection{Synth-aided margin separation module}
\label{subsec: SMS_module}

\begin{figure}[t]
  \centering
  \includegraphics[width=\linewidth]{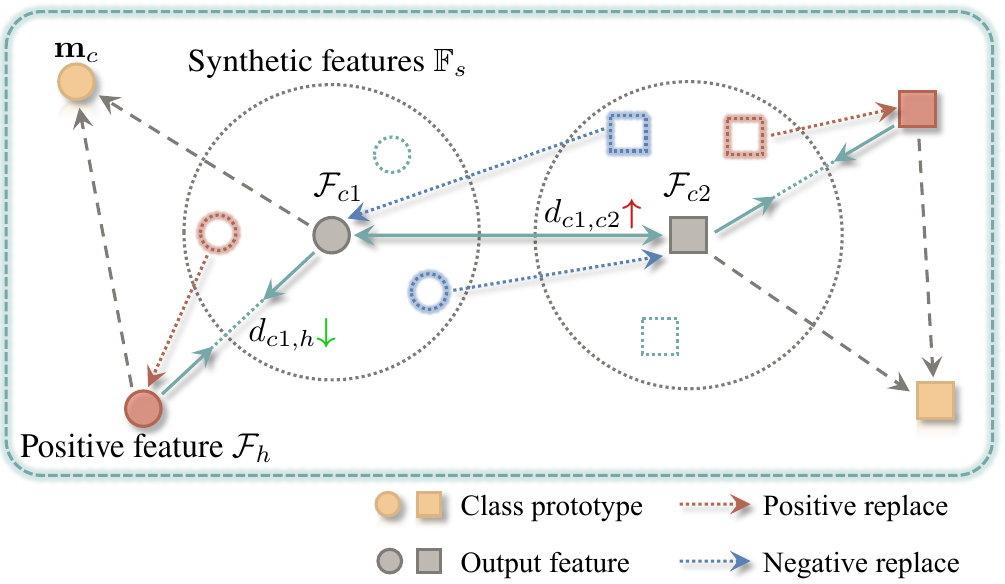}
    \vspace{-0.55cm}
   \caption{
   Illustration of our proposed synth-aided margin separation (SMS) module.
   The SMS module regularizes the feature space with a triplet loss, which employs the pseudo-features to help form the positive examples and negative examples. 
   }
   \label{fig: SMS_module}
   \vspace{-0.3cm}
\end{figure}

The synth-aided margin separation (SMS) module is designed to improve the compactness within each class and the separability between different classes. It achieves this by applying feature constraints to individual object features.
The key idea of SMS is illustrated in~\cref{fig: SMS_module}.
We denote the global feature of the input point cloud $\mathcal{P}_{c1}$ and its high-saliency part $\mathcal{P}_{h}$ as $\mathcal{F}_{c1}$ and $\mathcal{F}_h$, respectively.
To highlight the semantically-relevant local geometric structures in $\mathcal{P}_{c1}$, we design a dual alignment strategy comprising attraction and repulsion forces.
First, $\mathcal{P}_{h}$ is designated as a positive exemplar, and its feature $\mathcal{F}_h$ is aligned towards $\mathcal{F}_{c1}$ to consolidate intra-class similarity.
Concurrently, to enforce inter-class discrepancy, the negative exemplar $\mathcal{P}_{c2}$~($c1\neq c2$) is pushed away via its feature $\mathcal{F}_{c2}$ in the embedding space. 
By using both positive and negative exemplars, the feature constraint $\mathcal{L}_m$ is defined through a triplet loss framework to ensure that $d(\mathcal{F}_{c1}, \mathcal{F}_{h})$ is smaller than $d(\mathcal{F}_{c1}, \mathcal{F}_{c2})$.
\begin{equation}
\label{eq: triple_loss}
    \mathcal{L}_m = d(\mathcal{F}_{c1}, \mathcal{F}_{h}) - d(\mathcal{F}_{c1}, \mathcal{F}_{c2}) + \tau_m,
\end{equation}
where $d(\cdot)$ is the distance metric function.


Although $\mathcal{F}_{h}$ and $\mathcal{F}_{c2}$ are appropriate for constructing feature constraints, their performance may be limited by data scarcity due to the sparse distribution of feature samples.
To increase the availability of feature pairs, we generate synthetic features by adding Gaussian noise to $\mathcal{F}_{c1}$.
\begin{equation}
    \mathbb{F}_{s} = \mathcal{F}_{c1} + \mathcal{N}(0,\mathcal{W}),
\end{equation}
where $\mathcal{W}= \left\{ w_1, \cdots, w_n \right\}$ is a list of weights for Gaussian noise.
Then, we filter $\mathbb{F}_{s}$ to retain only the samples whose predicted results match the ground-truth labels.
This step enhances the reliability of synthetic features. 
From these filtered samples, one is randomly selected to serve as the pseudo feature.
This pseudo feature is then utilized for either random positive or negative replacements by substituting $\mathcal{F}_h$ or $\mathcal{F}_{c2}$, resulting in the creation of new positive exemplars $\mathcal{F}_p$ or negative exemplars $\mathcal{F}_n$.
In the case of positive replacement, the distance $d(\mathcal{F}_{c1}, \mathcal{F}_{h})$ is adjusted to accommodate a broader variability in $d(\mathcal{F}_{c1}, \mathcal{F}_{p})$, allowing for greater flexibility compared to the original distance.
Conversely, negative replacement is designed to enhance the robustness of object features by introducing disturbances directly into the feature space, applying force on $\mathcal{F}_{c1}$ with $d(\mathcal{F}_{c1}, \mathcal{F}_{n})$ as highlighted in~\cref{eq: triple_loss}.
It's important to note that only one of the positive or negative replacements will be executed for each input feature.
Finally, the $\mathcal{L}_m$ is refined with a weighted triplet loss:
\begin{equation}
\label{eq: feat_const_loss}
    \mathcal{L}_m = \rho d(\mathcal{F}_{c1}, \mathcal{F}_{p}) - \eta d(\mathcal{F}_{c1}, \mathcal{F}_{n}) + \tau_m,
\end{equation}
where $\rho$ and $\eta$ are the balance weights and $\tau_m$ is the margin threshold.
Given that the positive exemplar pairs are already constrained by the classification loss, we set $\rho < \eta$ to prioritize the enhancement of $d(\mathcal{F}_{c1}, \mathcal{F}_{n})$.

\subsection{Loss function}
\label{subsec: loss_fun}

The total loss of SASep integrates four components: (1) the \textbf{primary classification loss} $\mathcal{L}_{cls}$ for raw inputs $\mathcal{P}_{in}$, (2) the \textbf{high-saliency loss} $\mathcal{L}_{h}$ for supervising TSD-generated high-saliency parts $\mathcal{P}_{h}$, (3) the \textbf{synthesis loss} $\mathcal{L}_{s}$ (\cref{eq: gss_loss}) for synthetic samples from the GSS module, and (4) the \textbf{feature margin loss} $\mathcal{L}_m$ (\cref{eq: feat_const_loss}).
For $\mathcal{P}_{in}$ of class $c$, $\mathcal{L}_{cls}$ applies cross-entropy to the model output $\mathcal{Y}_{in}$:
\begin{equation}
\label{eq: cls_loss}
    \mathcal{L}_{cls} = - \log \frac{e^{\mathcal{Y}_{in}^c}}{\sum_{i=1}^{C+1} e^{\mathcal{Y}_{in}^i}},
\end{equation}
where $\mathcal{Y}_{in}^i$ is the cosine similarity between the encoded feature $\mathcal{F}_{in} \in \mathbb{R}^d$ of $\mathcal{P}_{in}$ and learnable prototypes $\mathcal{M} = \{ \mathbf{m}_1, \cdots, \mathbf{m}_C, \mathbf{m}_{C+1} \} \in \mathbb{R}^{(C+1)\times d}$, with $\mathbf{m}_{C+1}$ for unknown classes.
Hence, $\mathcal{Y}_{in}^i$ is defined as follows:
\begin{equation}
\label{eq: cosine_sim}
    \mathcal{Y}_{in}^i = g(\mathcal{F}_{in}; \mathbf{m}_i) = \frac{ \mathcal{F}_{in} \cdot \mathbf{m}_i }{ \lVert \mathcal{F}_{in} \rVert \lVert \mathbf{m}_i \rVert },
\end{equation}
where $\lVert \cdot \rVert$ denotes the $L2$-norm.
Note that the prototypes $\mathcal{M}$ are initialized randomly and optimized during training.
As aforementioned in~\cref{subsec: overview}, we consider TSD-generated high-saliency parts $\mathcal{P}_{h}$ as new instances belonging to the same class as $\mathcal{P}_{in}$ and directly feed $\mathcal{P}_{h}$ into the model.
Consequently, we compute $\mathcal{L}_{h}$ for $\mathcal{P}_{h}$ using the same formulation outlined in~\cref{eq: cls_loss}.
The overall open-set classification loss $\mathcal{L}$ is thus defined as:
\begin{equation}
\label{eq: total_loss}
    \mathcal{L} = \mathcal{L}_{cls} + \alpha \mathcal{L}_{h} + \beta \mathcal{L}_{s} + \gamma \mathcal{L}_m,
\end{equation}
where $\alpha$, $\beta$ and $\gamma$ are balance parameters.
\section{Experiments}
\label{sec:experiment}

\begin{table*}[!t]
    \centering
    \caption{The results of unknown sample identification task on the synthetic, synthetic-to-real, and real-to-real benchmarks. Each column shows the average results of each benchmark's subgroups. The best results are in bold in each metric.}
    \label{tab: unknown_identification}
    \setlength{\tabcolsep}{3pt}{
    \resizebox{\textwidth}{!}{
    \begin{tabular}{l|cc|cc|cc|cc|cc|cc}
    \toprule[1.1pt]
        \multirow{3}{*}{~} & \multicolumn{6}{c|}{DGCNN} & \multicolumn{6}{c}{PointNet++} \\  \cline{2-13} 
        & \multicolumn{2}{c|}{Synthetic} & \multicolumn{2}{c|}{Synthetic-to-Real} & \multicolumn{2}{c|}{Real-to-Real} & \multicolumn{2}{c|}{Synthetic} & \multicolumn{2}{c|}{Synthetic-to-Real} & \multicolumn{2}{c}{Real-to-Real} \\ \cline{2-13}
        Methods & AUROC$\uparrow$ & FPR95$\downarrow$ & AUROC$\uparrow$ & FPR95$\downarrow$ & AUROC$\uparrow$ & FPR95$\downarrow$ &  AUROC$\uparrow$ & FPR95$\downarrow$ & AUROC$\uparrow$ & FPR95$\downarrow$ & AUROC$\uparrow$ & FPR95$\downarrow$\\ \midrule[0.75pt]
        MSP~\cite{hendrycks2017a} & 85.2 & 63.2 & 66.7 & 90.6 & 70.8 & 82.8 & 81.3 & 71.0 & 75.6 & 83.2 & 80.8 & 77.2 \\ \hline
        MLS~\cite{pmlr-v162-hendrycks22a} & 86.2 & 51.8 & 65.7 & 90.5 & 72.6 & 77.3 & 81.9 & 61.8 & 74.8 & 81.7 & 83.0 & 67.4 \\ \hline
        ODIN~\cite{liang2018enhancing} & 86.3 & 51.3 & 65.7 & 90.6 & 72.6 & 77.3 & 80.5 & 64.3 & 76.0 & 80.8 & 83.3 & 65.5 \\ \hline
        Energy~\cite{Wang_2021_ICCV} & 86.2 & 51.7 & 65.6 & 90.8 & 72.6 & 78.1 & 82.0 & 61.8 & 74.8 & 82.4 & 82.6 & 67.3 \\ \hline
        GradNorm~\cite{NEURIPS2021_063e26c6} & 72.9 & 76.6 & 63.4 & 91.5 & 67.2 & 84.5 & 62.6 & 83.3 & 73.0 & 83.2 & 80.4 & 67.6 \\ \hline
        ReAct~\cite{NEURIPS2021_01894d6f} & 88.4 & 43.9 & 65.6 & 90.5 & 75.1 & 75.3 & 86.1 & 52.9 & 74.6 & 81.4 & 82.7 & 67.4 \\ \midrule[0.75pt]
        VAE~\cite{9506795} & 77.0 & 64.2 & 63.3 & 84.6 & - & - & - & - & - & - & - & -\\ \hline
        NF & 88.5 & 46.7 & 71.3 & 82.3 & 70.2 & 82.1 & 81.2 & 66.7 & 76.4 & 84.3 & 81.4 & 65.8 \\ \midrule[0.75pt]
        OE+mixup~\cite{Lee_2021_CVPR} & 85.2 & 56.5 & 65.3 & 90.8 & 69.8 & 85.0 & 80.1 & 69.9 & 65.7 & 91.6 & 69.0 & 86.6 \\
        PointCaM~\cite{hong2022pointcam} & - & - & 71.1 & 86.6 & - & - & - & - & - & - & - & - \\ \midrule[0.75pt]
        ARPL+CS~\cite{9521769} & 84.4 & 73.6 & 67.1 & 89.8 & - & - & 80.3 & 67.8 & 75.4 & 82.1 & - & - \\ \hline
        Cosine proto~\cite{Fontanel_2021_CVPR} & 86.5 & 48.9 & 57.9 & 91.0 & 78.0 & 68.2 & 86.9 & 55.7 & 78.2 & 76.1 & 83.6 & 64.9\\ \hline
        CE (L2) & 89.1 & 43.6 & 66.1 & 89.2 & 74.9 & 74.4 & 88.6 & 44.2 & 77.7 & 82.3 & 79.1 & 75.8 \\ \hline
        SupCon~\cite{NEURIPS2020_d89a66c7} & 84.3 & 64.5 & - & - & - & - & 83.2 & 62.9 & - & - & - & - \\ \hline
        SubArcFace~\cite{10.1007/978-3-030-58621-8_43} & 89.3 & 48.0 & 71.6 & 86.7 & 76.7 & 72.9 & 83.7 & 58.1 & 76.9 & 83.8 & 79.9 & 73.2 \\ \midrule[0.75pt]
        \textbf{Ours} & \textbf{92.5} & \textbf{28.7} & \textbf{77.9} & \textbf{75.4} & \textbf{82.0} & \textbf{62.3} & \textbf{92.8} & \textbf{29.1} & \textbf{79.7} & \textbf{67.5} & \textbf{84.8} & \textbf{55.0}\\ \bottomrule[1.1pt]
    \end{tabular}
    }
    }
\end{table*}

\subsection{Datasets}

We conduct experiments using the benchmarks proposed by 3DOS~\cite{NEURIPS2022_85b6841e}, which are based on three widely used 3D object datasets: ShapeNetCore~\cite{chang2015shapenet}, ModelNet40~\cite{Wu_2015_CVPR}, and ScanObjectNN~\cite{Uy_2019_ICCV}.
The ShapeNetCore dataset is a large synthetic 3D dataset that contains 51,127 3D mesh instances across 55 object categories.
Approximately 70\% of the data is allocated for training, 10\% for validation, and 20\% for testing.
The ModelNet40 dataset consists of 12,311 3D CAD-generated meshes from 40 categories.
This dataset is divided into an official training set of 9,843 samples and a test set of 2,468 samples.
The corresponding point cloud data points are uniformly sampled from the surfaces of the CAD meshes and scaled to fit with a unit sphere.
ScanObjectNN is a real-world dataset comprising 2,902 3D scans across 15 categories.
It presents significant challenges due to artifacts, such as background clutter, occlusions, missing parts, and non-uniform density.

We follow the original experimental settings for open-set 3D recognition defined by 3DOS benchmarks, which include three tracks designed to evaluate the performance under different conditions of semantic and domain shifts.
In the synthetic benchmark, the ShapeNetCore dataset is divided into three distinct category subgroups, designated as SN1, SN2 and SN3, each containing 18 classes with no overlap.
Each group rotates as the set of known classes, while the other two are treated as unknown classes, creating three unique experimental configurations.
Similarly, the real-to-real benchmark follows the ScanObjectNN dataset, which is also divided into three subgroups (RR1, RR2, and RR3) to create three corresponding experimental configurations under real-scanned conditions.
The synthetic-to-real benchmark establishes cross-domain scenarios (SR1, SR2), wherein synthetic samples from ModelNet40 serve as the known class set, while real scans from ScanObjectNN are designated as the unknown class set.
Given the significant domain shifts between the known and unknown categories, this synthetic-to-real benchmark presents greater challenges compared to the other two benchmarks.

\subsection{Evaluation metrics}

Open-set 3D recognition includes two main tasks: closed-set classification and identification of unknown samples.
For the closed-set classification task, we utilize accuracy (ACC) and mean accuracy (mACC) to evaluate the model's performance in correctly classifying known samples.
To assess performance in identifying unknown samples, we utilize AUROC and FPR95 metrics.
The AUROC (Area Under the Receiver Operating Characteristic Curve) measures a model's ability to differentiate between known and unknown samples.
The ROC curve plots the true positive rate against the false positive rate across various thresholds to reject unknown samples.
A higher AUROC value indicates better performance in identifying unknown samples.
FPR95(False Positive Rate at 95\% True Positive Rate) measures the rate of false positives while keeping the true positive rate fixed at 95\%.
This metric offers insights into the model's capacity in reducing misclassifications, particularly in preventing the incorrect identification of unknown samples as known.

\subsection{Implementation details}

Our proposed SASep method is implemented using the PyTorch framework and evaluated with DGCNN~\cite{wang2019dynamic} and PointNet++~\cite{qi2017pointnet++}.
To simplify the training process, saliency maps for each object were generated using Grad-CAM~\cite{Selvaraju_2017_ICCV} with the pre-trained closed-set backbones.
All models are trained on 4 NVIDIA RTX 3090 GPUs for 200 epochs, with a total batch size of 120.
For both DGCNN and PointNet++, we adopt the Adam optimizer and CosLR scheduler with an initial learning rate set to 0.001.
The balance parameters $\alpha$, $\beta$, and $\gamma$ in~\cref{eq: total_loss} are assigned to 0.1, 0.01, and 0.3, respectively.
In~\cref{eq: feat_const_loss}, the weights $\rho$ and $\eta$ are set to 0.01 and 1.0, respectively, with $\tau_m$ fixed at 10.

\subsection{Comparison on unknown sample identification}

\begin{figure*}[!t]
    \centering
    \includegraphics[width=1\textwidth]{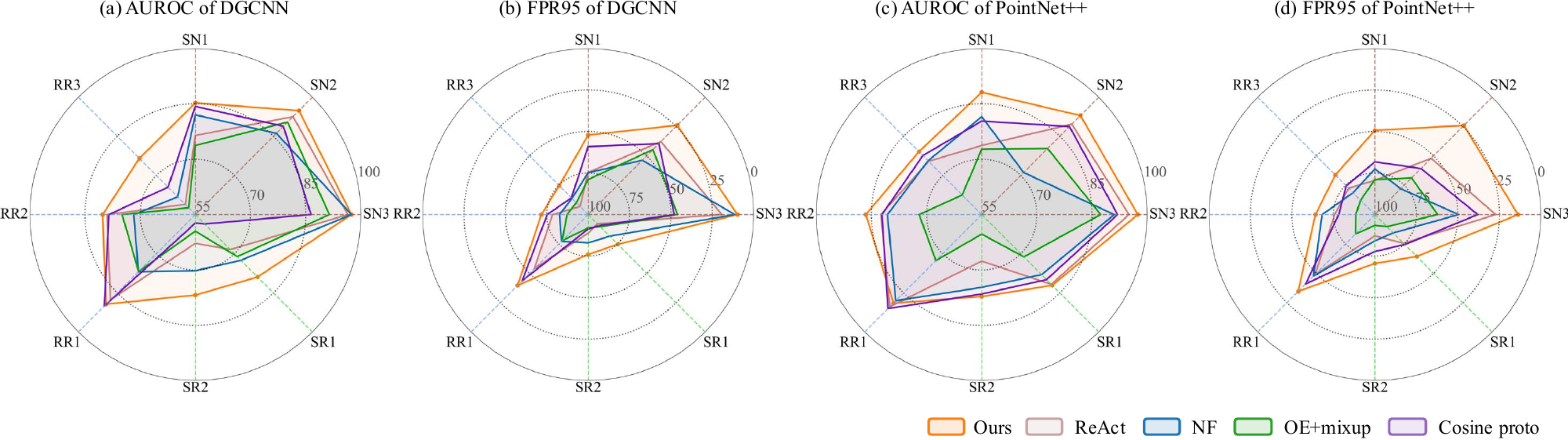}

    \caption{Comparisons on all the subgroups of synthetic, synthetic-to-real, real-to-real benchmarks. The AUROC and FPR95 metrics on the DGCNN and PointNet++ are drawn in the radar chart (a)-(d). The results of our proposed SASep are highlighted in orange.}
    \label{fig: radar_metric}
\end{figure*}

We compare our proposed SASep method with various existing open-set recognition methods, including both 2D and 3D techniques, across all tracks on synthetic, synth-to-real, and real-to-real benchmarks.
The results of each benchmark's subgroups are aggregated to provide a comprehensive evaluation.
As shown in~\cref{tab: unknown_identification}, our proposed method consistently demonstrates the highest AUROC scores and the lowest FPR95 scores across all benchmarks and backbones.
Specifically, SASep surpasses the best-performing methods in each benchmark with AUROC and FPR95 improvements of at least 2\% and 6\%, respectively.
Notably, SASep achieves the highest improvements in the AUROC and FPR95 metrics on synthetic-to-real and synthetic benchmarks, with gains of 6.3\% and 15.1\%, respectively.
Furthermore, SASep demonstrates higher improvements in the synthetic benchmark compared to the other benchmarks.
We believe this is due to the characteristics of the synthetic data, which has more complete geometric structures and less real-world noise.
This allows SASep to effectively capture the class-relevant structures, leading to improved performance.
Apart from the synthetic benchmark, the improvements observed in the synthetic-to-real and real-to-real benchmarks further highlight the robustness of our method in addressing the challenges presented by the cross-domain gap between known and unknown categories.

To provide a more intuitive comparison of our method with other representative methods, we visualize the AUROC and FPR95 metrics for both DGCNN and PointNet++ across all subgroups of each benchmark using radar charts, as shown in~\cref{fig: radar_metric}.
Clearly, our proposed method (highlighted in orange) outperforms other methods in nearly all experiments, demonstrating its superior performance.
Please refer to the Supplemental Material for more detailed experimental results.

\subsection{Comparison on closed-set classification}

We evaluate the closed-set classification performance of our method against baseline closed-set models with DGCNN and PointNet++ backbones. 
The closed-set models are trained and tested solely on known classes.
Although our method further performs the unknown sample identification task, we only compare the classification accuracy.
\cref{tab: closed-set} shows the accuracy (ACC) of each method across three subgroups: SN1, SR1, and RR1.
For the DGCNN backbone, our method achieves improved performance in SR1 and RR1, reaching 81.3\% and 94\%, respectively, compared to the closed-set model’s 72.1\% and 91.5\%, with a slight decrease in SN1.
The improvements observed in SR1 and RR1 demonstrate the enhanced robustness of our method when applied to real-world scan data.
This can be attributed to our design, which focuses on capturing semantically important geometric structures, thereby improving the model’s resilience to real-world scenarios.
For the PointNet++ backbone, our method achieves results comparable to the closed-set model, with only a slight decrease in performance.
This may be due to PointNet++'s stronger capacity for feature extraction using hierarchical network structures.
Overall, these results suggest that our method not only competes with closed-set models in terms of accuracy but also provides increased robustness against real-world challenges.

\begin{table}[!t]
    \centering
    \caption{Results of closed-set classification task on the subgroup (SN1, SR1, RR1) of synthetic, synthetic-to-real, and real-to-real benchmarks. The ACC metric is used as the evaluation metric.}
    \label{tab: closed-set}
    \resizebox{0.85\linewidth}{!}{
    \begin{tabular}{l|c|c|c|c|c|c}
    \toprule[1.1pt]
        \multirow{2}{*}{~} & \multicolumn{3}{c|}{DGCNN} & \multicolumn{3}{c}{PointNet++} \\ \cline{2-7} 
        Methods & SN1 & SR1 & RR1 & SN1 & SR1 & RR1 \\ \midrule[0.75pt]
        closed-set & \textbf{84.7} & 72.1 & 91.5 & \textbf{85.9} & 80.0 & \textbf{95.6} \\ \hline
        Our & 83.3 & \textbf{81.3} & \textbf{94.0} & 85.4 & \textbf{80.0} & 94.8 \\ \bottomrule[1.1pt]
    \end{tabular}
    }
\end{table}

\subsection{Ablation study}

To evaluate the contribution of each component in our proposed method, we conduct ablation studies on SN2 of the synthetic benchmark with DGCNN backbone.
As shown in~\cref{tab: ablation}, we report the AUROC, FPR95, ACC, and mACC metrics across various configurations.

When all modules (TSD, GSS, SMS) are enabled, the model achieves its best performance, with an AUROC of 94.8, the lowest FPR95 at 24.0, and the highest ACC and mACC at 90.5 and 79.8, respectively.
This indicates that the full combination maximizes performance for both known class classification and unknown class identification tasks.

Removing each module clarifies specific strengths.
Excluding TSD (with only GSS and SMS active) leads to a decreased AUROC and mACC, demonstrating TSD's significant role in both tasks of open-set learning.
When GSS is removed, the AUROC slightly increases, but the FPR95 drops to 29.1, indicating GSS's key role in controlling false positives.
Without SMS, all the metrics are slightly decreased, suggesting SMS's impact on overall accuracy.

Individual module analysis highlights TSD's strength, achieving an AUROC of 94.2 and the lowest FPR95 of 25.9.
GSS alone enhances mACC but raises FPR95 to 30.6, while SMS alone has the highest FPR95, indicating its lower effectiveness in isolation.

\begin{table}[!t]
    \centering
    \caption{Comparing the contribution of major components in our proposed SASep on SN2 of synthetic benchmark using DGCNN.}
    \label{tab: ablation}
    \resizebox{\linewidth}{!}{
    \begin{tabular}{ccc|cccc}
    \toprule[1.1pt]
        TSD & GSS & SMS & AUROC$\uparrow$ & FPR95$\downarrow$ & ACC$\uparrow$ &  mACC$\uparrow$ \\ \midrule[0.75pt]
        ~ & \checkmark & \checkmark & 93.7 & 28.3 & 88.3 & 74.0 \\ \hline
        \checkmark & ~ & \checkmark & 94.1 & 29.1 & 89.9 & 78.4 \\ \hline
        \checkmark & \checkmark & ~ & 93.8 & 26.9 & 89.2 & 78.5 \\ \midrule[0.75pt]
        \checkmark & ~ & ~ & 94.2 & 25.9 & 90.0 & 77.6 \\ \hline
        ~ & \checkmark & ~ & 93.6 & 30.6 & 88.6 & 79.0 \\ \hline
        ~ & ~ & \checkmark & 93.0 & 36.1 & 89.6 & 75.9 \\ \midrule[0.75pt]
        \checkmark & \checkmark & \checkmark & \textbf{94.8} & \textbf{24.0} & \textbf{90.5} & \textbf{79.8} \\ \bottomrule[1.1pt]
    \end{tabular}
    }
\end{table}



\section{Conclusion}
\label{sec:conclusion}

In this work, we propose a novel open-set point cloud learning method, SASep, which identifies unknown classes by emphasizing the importance of utilizing the geometric structures of objects' different parts, rather than treating them as a whole.
Specifically, we design a tunable semantic decomposition module to decompose objects into the semantically important and unimportant parts for classification.
Then, we introduce a geometric synthesis strategy and a synth-aided margin separation module to leverage these decomposed parts, improving feature learning for known classes and enhancing class distinction. 
Experimental results show that SASep significantly outperforms the state-of-the-art method in various datasets.
In scenarios with severe corruption, our method may be limited due to challenges in accurately capturing semantics, motivating us to explore a more robust method for semantic decomposition.
We hope SASep offers some new insights into this underexplored area.

\section*{Acknowledgements}

This work was supported by the National Key Research and Development Program for Young Scientists of China (2024YFB3310100), the China National Natural Science Foundation No.~62202182, the Guangdong Basic and Applied Basic Research Foundation 2024A1515010224, the National Key Research and Development Program of China under Grant 2023YFB4503400, and the China National Natural Science Foundation Nos.~62176101, 62276109, 62450064, and~62322205.
{
    \small
    \bibliographystyle{ieeenat_fullname}
    \bibliography{main}

\begin{thebibliography}{46}
\providecommand{\natexlab}[1]{#1}
\providecommand{\url}[1]{\texttt{#1}}
\expandafter\ifx\csname urlstyle\endcsname\relax
  \providecommand{\doi}[1]{doi: #1}\else
  \providecommand{\doi}{doi: \begingroup \urlstyle{rm}\Url}\fi

\bibitem[Alliegro et~al.(2022)Alliegro, Cappio~Borlino, and
  Tommasi]{NEURIPS2022_85b6841e}
Antonio Alliegro, Francesco Cappio~Borlino, and Tatiana Tommasi.
\newblock 3dos: Towards 3d open set learning - benchmarking and understanding
  semantic novelty detection on point clouds.
\newblock In \emph{NeurIPS}, pages 21228--21240. Curran Associates, Inc., 2022.

\bibitem[Bhardwaj et~al.(2021)Bhardwaj, Pimpale, Kumar, and
  Banerjee]{BHARDWAJ2021172}
Ayush Bhardwaj, Sakshee Pimpale, Saurabh Kumar, and Biplab Banerjee.
\newblock Empowering knowledge distillation via open set recognition for robust
  3d point cloud classification.
\newblock \emph{Pattern Recognition Letters}, 151:\penalty0 172--179, 2021.

\bibitem[Chang et~al.(2015)Chang, Funkhouser, Guibas, Hanrahan, Huang, Li,
  Savarese, Savva, Song, Su, et~al.]{chang2015shapenet}
Angel~X Chang, Thomas Funkhouser, Leonidas Guibas, Pat Hanrahan, Qixing Huang,
  Zimo Li, Silvio Savarese, Manolis Savva, Shuran Song, Hao Su, et~al.
\newblock Shapenet: An information-rich 3d model repository.
\newblock \emph{arXiv preprint arXiv:1512.03012}, 2015.

\bibitem[Chen et~al.(2022)Chen, Peng, Wang, and Tian]{9521769}
Guangyao Chen, Peixi Peng, Xiangqian Wang, and Yonghong Tian.
\newblock { Adversarial Reciprocal Points Learning for Open Set Recognition }.
\newblock \emph{IEEE TPAMI}, 44\penalty0 (11):\penalty0 8065--8081, 2022.

\bibitem[Chen et~al.(2017)Chen, Ma, Wan, Li, and Xia]{Chen_2017_CVPR}
Xiaozhi Chen, Huimin Ma, Ji Wan, Bo Li, and Tian Xia.
\newblock Multi-view 3d object detection network for autonomous driving.
\newblock In \emph{CVPR}, 2017.

\bibitem[Choy et~al.(2019)Choy, Gwak, and Savarese]{Choy_2019_CVPR}
Christopher Choy, JunYoung Gwak, and Silvio Savarese.
\newblock 4d spatio-temporal convnets: Minkowski convolutional neural networks.
\newblock In \emph{CVPR}, 2019.

\bibitem[Deng et~al.(2020)Deng, Guo, Liu, Gong, and
  Zafeiriou]{10.1007/978-3-030-58621-8_43}
Jiankang Deng, Jia Guo, Tongliang Liu, Mingming Gong, and Stefanos Zafeiriou.
\newblock Sub-center arcface: Boosting face recognition by large-scale noisy
  web faces.
\newblock In \emph{ECCV}, pages 741--757, Cham, 2020. Springer International
  Publishing.

\bibitem[Du et~al.(2022{\natexlab{a}})Du, Wang, Cai, and Li]{du2022towards}
Xuefeng Du, Zhaoning Wang, Mu Cai, and Sharon Li.
\newblock Towards unknown-aware learning with virtual outlier synthesis.
\newblock In \emph{ICLR}, 2022{\natexlab{a}}.

\bibitem[Du et~al.(2022{\natexlab{b}})Du, Wang, Cai, and Li]{du2022vos}
Xuefeng Du, Zhaoning Wang, Mu Cai, and Yixuan Li.
\newblock Vos: Learning what you don’t know by virtual outlier synthesis.
\newblock \emph{ICLR}, 2022{\natexlab{b}}.

\bibitem[Fontanel et~al.(2021)Fontanel, Cermelli, Mancini, and
  Caputo]{Fontanel_2021_CVPR}
Dario Fontanel, Fabio Cermelli, Massimiliano Mancini, and Barbara Caputo.
\newblock Detecting anomalies in semantic segmentation with prototypes.
\newblock In \emph{Proceedings of the IEEE/CVF Conference on Computer Vision
  and Pattern Recognition (CVPR) Workshops}, pages 113--121, 2021.

\bibitem[Graham et~al.(2018)Graham, Engelcke, and van~der
  Maaten]{Graham_2018_CVPR}
Benjamin Graham, Martin Engelcke, and Laurens van~der Maaten.
\newblock 3d semantic segmentation with submanifold sparse convolutional
  networks.
\newblock In \emph{CVPR}, 2018.

\bibitem[Han et~al.(2024)Han, Tang, Wang, and Li]{10.1145/3664647.3681173}
Xu Han, Yuan Tang, Zhaoxuan Wang, and Xianzhi Li.
\newblock Mamba3d: Enhancing local features for 3d point cloud analysis via
  state space model.
\newblock In \emph{ACM MM}, page 4995–5004, New York, NY, USA, 2024.
  Association for Computing Machinery.

\bibitem[Hendrycks and Gimpel(2017)]{hendrycks2017a}
Dan Hendrycks and Kevin Gimpel.
\newblock A baseline for detecting misclassified and out-of-distribution
  examples in neural networks.
\newblock In \emph{ICLR}, 2017.

\bibitem[Hendrycks et~al.(2022)Hendrycks, Basart, Mazeika, Zou, Kwon,
  Mostajabi, Steinhardt, and Song]{pmlr-v162-hendrycks22a}
Dan Hendrycks, Steven Basart, Mantas Mazeika, Andy Zou, Joseph Kwon,
  Mohammadreza Mostajabi, Jacob Steinhardt, and Dawn Song.
\newblock Scaling out-of-distribution detection for real-world settings.
\newblock In \emph{Proceedings of the 39th International Conference on Machine
  Learning}, pages 8759--8773. PMLR, 2022.

\bibitem[Hong et~al.(2022)Hong, Qiu, Li, Anwar, Harandi, Barnes, and
  Petersson]{hong2022pointcam}
Jie Hong, Shi Qiu, Weihao Li, Saeed Anwar, Mehrtash Harandi, Nick Barnes, and
  Lars Petersson.
\newblock Pointcam: Cut-and-mix for open-set point cloud learning.
\newblock \emph{arXiv preprint arXiv:2212.02011}, 2022.

\bibitem[Huang et~al.(2021)Huang, Geng, and Li]{NEURIPS2021_063e26c6}
Rui Huang, Andrew Geng, and Yixuan Li.
\newblock On the importance of gradients for detecting distributional shifts in
  the wild.
\newblock In \emph{NeurIPS}, pages 677--689. Curran Associates, Inc., 2021.

\bibitem[Jo et~al.(2018)Jo, Kim, Kang, Kim, and Choi]{8461700}
Inhyuk Jo, Jungtaek Kim, Hyohyeong Kang, Yong-Deok Kim, and Seungjin Choi.
\newblock Open set recognition by regularising classifier with fake data
  generated by generative adversarial networks.
\newblock In \emph{2018 IEEE International Conference on Acoustics, Speech and
  Signal Processing (ICASSP)}, pages 2686--2690, 2018.

\bibitem[Khosla et~al.(2020)Khosla, Teterwak, Wang, Sarna, Tian, Isola,
  Maschinot, Liu, and Krishnan]{NEURIPS2020_d89a66c7}
Prannay Khosla, Piotr Teterwak, Chen Wang, Aaron Sarna, Yonglong Tian, Phillip
  Isola, Aaron Maschinot, Ce Liu, and Dilip Krishnan.
\newblock Supervised contrastive learning.
\newblock In \emph{NeurIPS}, pages 18661--18673. Curran Associates, Inc., 2020.

\bibitem[Kong and Ramanan(2021)]{Kong_2021_ICCV}
Shu Kong and Deva Ramanan.
\newblock Opengan: Open-set recognition via open data generation.
\newblock In \emph{ICCV}, pages 813--822, 2021.

\bibitem[Lai et~al.(2022)Lai, Liu, Jiang, Wang, Zhao, Liu, Qi, and
  Jia]{Lai_2022_CVPR}
Xin Lai, Jianhui Liu, Li Jiang, Liwei Wang, Hengshuang Zhao, Shu Liu, Xiaojuan
  Qi, and Jiaya Jia.
\newblock Stratified transformer for 3d point cloud segmentation.
\newblock In \emph{CVPR}, pages 8500--8509, 2022.

\bibitem[Lang et~al.(2019)Lang, Vora, Caesar, Zhou, Yang, and
  Beijbom]{Lang_2019_CVPR}
Alex~H. Lang, Sourabh Vora, Holger Caesar, Lubing Zhou, Jiong Yang, and Oscar
  Beijbom.
\newblock Pointpillars: Fast encoders for object detection from point clouds.
\newblock In \emph{CVPR}, 2019.

\bibitem[Lee et~al.(2021)Lee, Lee, Lee, Lee, Lee, Woo, and Lee]{Lee_2021_CVPR}
Dogyoon Lee, Jaeha Lee, Junhyeop Lee, Hyeongmin Lee, Minhyeok Lee, Sungmin Woo,
  and Sangyoun Lee.
\newblock Regularization strategy for point cloud via rigidly mixed sample.
\newblock In \emph{CVPR}, pages 15900--15909, 2021.

\bibitem[Lei et~al.(2019)Lei, Akhtar, and Mian]{Lei_2019_CVPR}
Huan Lei, Naveed Akhtar, and Ajmal Mian.
\newblock Octree guided cnn with spherical kernels for 3d point clouds.
\newblock In \emph{CVPR}, 2019.

\bibitem[Liang et~al.(2018)Liang, Li, and Srikant]{liang2018enhancing}
Shiyu Liang, Yixuan Li, and R. Srikant.
\newblock Enhancing the reliability of out-of-distribution image detection in
  neural networks.
\newblock In \emph{ICLR}, 2018.

\bibitem[Liu et~al.(2023)Liu, Zhou, Xu, and Wang]{Liu_2023_CVPR}
Zhikang Liu, Yiming Zhou, Yuansheng Xu, and Zilei Wang.
\newblock Simplenet: A simple network for image anomaly detection and
  localization.
\newblock In \emph{CVPR}, pages 20402--20411, 2023.

\bibitem[Ma et~al.(2022)Ma, Qin, You, Ran, and Fu]{ma2022rethinking}
Xu Ma, Can Qin, Haoxuan You, Haoxi Ran, and Yun Fu.
\newblock Rethinking network design and local geometry in point cloud: A simple
  residual {MLP} framework.
\newblock In \emph{ICLR}, 2022.

\bibitem[Masuda et~al.(2021)Masuda, Hachiuma, Fujii, Saito, and
  Sekikawa]{9506795}
Mana Masuda, Ryo Hachiuma, Ryo Fujii, Hideo Saito, and Yusuke Sekikawa.
\newblock Toward unsupervised 3d point cloud anomaly detection using
  variational autoencoder.
\newblock In \emph{ICIP}, pages 3118--3122, 2021.

\bibitem[Maturana and Scherer(2015)]{7353481}
Daniel Maturana and Sebastian Scherer.
\newblock Voxnet: A 3d convolutional neural network for real-time object
  recognition.
\newblock In \emph{2015 IEEE/RSJ International Conference on Intelligent Robots
  and Systems (IROS)}, pages 922--928, 2015.

\bibitem[Mehra et~al.(2010)Mehra, Tripathi, Sheffer, and
  Mitra]{mehra2010visibility}
Ravish Mehra, Pushkar Tripathi, Alla Sheffer, and Niloy~J Mitra.
\newblock Visibility of noisy point cloud data.
\newblock \emph{Computers \& Graphics}, 34\penalty0 (3):\penalty0 219--230,
  2010.

\bibitem[Neal et~al.(2018)Neal, Olson, Fern, Wong, and Li]{Neal_2018_ECCV}
Lawrence Neal, Matthew Olson, Xiaoli Fern, Weng-Keen Wong, and Fuxin Li.
\newblock Open set learning with counterfactual images.
\newblock In \emph{ECCV}, 2018.

\bibitem[Qi et~al.(2017{\natexlab{a}})Qi, Su, Mo, and Guibas]{Qi_2017_CVPR}
Charles~R. Qi, Hao Su, Kaichun Mo, and Leonidas~J. Guibas.
\newblock Pointnet: Deep learning on point sets for 3d classification and
  segmentation.
\newblock In \emph{CVPR}, 2017{\natexlab{a}}.

\bibitem[Qi et~al.(2017{\natexlab{b}})Qi, Yi, Su, and Guibas]{qi2017pointnet++}
Charles~Ruizhongtai Qi, Li Yi, Hao Su, and Leonidas~J Guibas.
\newblock Pointnet++: Deep hierarchical feature learning on point sets in a
  metric space.
\newblock In \emph{NeurIPS}, 2017{\natexlab{b}}.

\bibitem[Rabino et~al.(2024)Rabino, Alliegro, and Tommasi]{10684168}
Paolo Rabino, Antonio Alliegro, and Tatiana Tommasi.
\newblock 3d semantic novelty detection via large-scale pre-trained models.
\newblock \emph{IEEE Access}, 12:\penalty0 135352--135361, 2024.

\bibitem[Selvaraju et~al.(2017)Selvaraju, Cogswell, Das, Vedantam, Parikh, and
  Batra]{Selvaraju_2017_ICCV}
Ramprasaath~R. Selvaraju, Michael Cogswell, Abhishek Das, Ramakrishna Vedantam,
  Devi Parikh, and Dhruv Batra.
\newblock Grad-cam: Visual explanations from deep networks via gradient-based
  localization.
\newblock In \emph{ICCV}, 2017.

\bibitem[Su et~al.(2015)Su, Maji, Kalogerakis, and
  Learned-Miller]{Su_2015_ICCV}
Hang Su, Subhransu Maji, Evangelos Kalogerakis, and Erik Learned-Miller.
\newblock Multi-view convolutional neural networks for 3d shape recognition.
\newblock In \emph{ICCV}, 2015.

\bibitem[Sun et~al.(2021)Sun, Guo, and Li]{NEURIPS2021_01894d6f}
Yiyou Sun, Chuan Guo, and Yixuan Li.
\newblock React: Out-of-distribution detection with rectified activations.
\newblock In \emph{NeurIPS}, pages 144--157. Curran Associates, Inc., 2021.

\bibitem[Tatarchenko et~al.(2018)Tatarchenko, Park, Koltun, and
  Zhou]{Tatarchenko_2018_CVPR}
Maxim Tatarchenko, Jaesik Park, Vladlen Koltun, and Qian-Yi Zhou.
\newblock Tangent convolutions for dense prediction in 3d.
\newblock In \emph{CVPR}, 2018.

\bibitem[Uy et~al.(2019)Uy, Pham, Hua, Nguyen, and Yeung]{Uy_2019_ICCV}
Mikaela~Angelina Uy, Quang-Hieu Pham, Binh-Son Hua, Thanh Nguyen, and Sai-Kit
  Yeung.
\newblock Revisiting point cloud classification: A new benchmark dataset and
  classification model on real-world data.
\newblock In \emph{ICCV}, 2019.

\bibitem[Wang et~al.(2017)Wang, Liu, Guo, Sun, and
  Tong]{10.1145/3072959.3073608}
Peng-Shuai Wang, Yang Liu, Yu-Xiao Guo, Chun-Yu Sun, and Xin Tong.
\newblock O-cnn: octree-based convolutional neural networks for 3d shape
  analysis.
\newblock \emph{ACM TOG}, 36\penalty0 (4), 2017.

\bibitem[Wang et~al.(2019)Wang, Sun, Liu, Sarma, Bronstein, and
  Solomon]{wang2019dynamic}
Yue Wang, Yongbin Sun, Ziwei Liu, Sanjay~E Sarma, Michael~M Bronstein, and
  Justin~M Solomon.
\newblock Dynamic graph cnn for learning on point clouds.
\newblock \emph{ACM Transactions on Graphics}, 38\penalty0 (5):\penalty0 1--12,
  2019.

\bibitem[Wang et~al.(2021)Wang, Li, Che, Zhou, Liu, and Li]{Wang_2021_ICCV}
Yezhen Wang, Bo Li, Tong Che, Kaiyang Zhou, Ziwei Liu, and Dongsheng Li.
\newblock Energy-based open-world uncertainty modeling for confidence
  calibration.
\newblock In \emph{ICCV}, pages 9302--9311, 2021.

\bibitem[Wang et~al.(2024)Wang, Mu, Zhu, and Hu]{Wang_Mu_Zhu_Hu_2024}
Yu Wang, Junxian Mu, Pengfei Zhu, and Qinghua Hu.
\newblock Exploring diverse representations for open set recognition.
\newblock \emph{AAAI}, 38\penalty0 (6):\penalty0 5731--5739, 2024.

\bibitem[Weng et~al.(2024)Weng, Xiao, Pan, and Jiang]{weng2024partcom}
Tingyu Weng, Jun Xiao, Hao Pan, and Haiyong Jiang.
\newblock Partcom: Part composition learning for 3d open-set recognition.
\newblock \emph{IJCV}, 132\penalty0 (4):\penalty0 1393--1416, 2024.

\bibitem[Wu et~al.(2024)Wu, Jiang, Wang, Liu, Liu, Qiao, Ouyang, He, and
  Zhao]{Wu_2024_CVPR}
Xiaoyang Wu, Li Jiang, Peng-Shuai Wang, Zhijian Liu, Xihui Liu, Yu Qiao, Wanli
  Ouyang, Tong He, and Hengshuang Zhao.
\newblock Point transformer v3: Simpler faster stronger.
\newblock In \emph{CVPR}, pages 4840--4851, 2024.

\bibitem[Wu et~al.(2015)Wu, Song, Khosla, Yu, Zhang, Tang, and
  Xiao]{Wu_2015_CVPR}
Zhirong Wu, Shuran Song, Aditya Khosla, Fisher Yu, Linguang Zhang, Xiaoou Tang,
  and Jianxiong Xiao.
\newblock 3d shapenets: A deep representation for volumetric shapes.
\newblock In \emph{CVPR}, 2015.

\bibitem[Zhao et~al.(2021)Zhao, Jiang, Jia, Torr, and Koltun]{Zhao_2021_ICCV}
Hengshuang Zhao, Li Jiang, Jiaya Jia, Philip~H.S. Torr, and Vladlen Koltun.
\newblock Point transformer.
\newblock In \emph{ICCV}, pages 16259--16268, 2021.

\end{thebibliography}
}


\end{document}